\documentclass[runningheads]{llncs}
\usepackage[T1]{fontenc}
\usepackage{graphicx}
\usepackage{amsmath,amssymb,amsfonts}
\usepackage{algorithmic}
\usepackage{graphicx}
\usepackage{textcomp}
\usepackage{xcolor}
\usepackage{float}
\restylefloat{table}
\usepackage{lscape}
 \usepackage[numbers]{natbib}
\usepackage{graphicx}
\usepackage{caption}
\usepackage{subcaption}
\usepackage{booktabs,makecell}
\usepackage{multirow}
\usepackage{pdflscape}
\usepackage{caption}
\usepackage{hyperref}
\usepackage{rotating}
\usepackage[export]{adjustbox}
\usepackage{comment}
\usepackage{amsmath}
\usepackage{xcolor}
\usepackage{tikz}
\usetikzlibrary{shapes,arrows}
\usepackage{placeins}
\usepackage{graphics}
\usepackage{xcolor}
\usepackage{mathrsfs}

\usepackage{algorithm}
\usepackage{algorithmic}

\usepackage{epstopdf}
\def\BibTeX{{\rm B\kern-.05em{\sc i\kern-.025em b}\kern-.08em
    T\kern-.1667em\lower.7ex\hbox{E}\kern-.125emX}}
\begin{document}
\title{One Class Restricted Kernel Machines}
\author{A. Quadir \and M. Sajid \and
M. Tanveer\thanks{ \noindent Corresponding Author}}
\authorrunning{A. Quadir et al.}
%
\institute{Indian Institute of Technology Indore, Simrol, Indore, India 
\email{\{mscphd2207141002,phd2101241003,mtanveer\}@iiti.ac.in}\\
}
\maketitle              
\begin{abstract}
Restricted kernel machines (RKMs) have demonstrated a significant impact in enhancing generalization ability in the field of machine learning. Recent studies have introduced various methods within the RKM framework, combining kernel functions with the least squares support vector machine (LSSVM) in a manner similar to the energy function of restricted boltzmann machines (RBM), such that a better performance can be achieved. However, RKM's efficacy can be compromised by the presence of outliers and other forms of contamination within the dataset. These anomalies can skew the learning process, leading to less accurate and reliable outcomes. To address this critical issue and to ensure the robustness of the model, we propose the novel one-class RKM (OCRKM). In the framework of OCRKM, we employ an energy function akin to that of the RBM, which integrates both visible and hidden variables in a nonprobabilistic setting. The formulation of the proposed OCRKM facilitates the seamless integration of one-class classification method with the RKM, enhancing its capability to detect outliers and anomalies effectively. The proposed OCRKM model is evaluated over UCI benchmark datasets. Experimental findings and statistical analyses consistently emphasize the superior generalization capabilities of the proposed OCRKM model over baseline models across all scenarios. The source code of the proposed OCRKM model is available at \url{https://github.com/mtanveer1/OCRKM}.
\keywords{One-class support vector machine \and Kernel methods \and Restricted kernel machine \and Restricted boltzmann machines.}
\end{abstract}

\section{Introduction}
Support vector machines (SVMs) \cite{cortes1995support} have become a powerful tool for addressing classification problems. SVM relies on statistical learning theory and the principle of maximizing the margin to find the best hyperplane that separates classes by solving a quadratic programming problem (QPP). SVM has been widely applied to various real-world problems, including remote sensing \cite{pal2005support}, electroencephalogram signal classification \cite{richhariya2018eeg}, diagnosis of significant memory concern \cite{Sajid2024smc}, feature extraction \cite{li2008joint}, and so on. SVM demonstrates superior performance by adhering to the structural risk minimization (SRM) principle, simultaneously minimizing the empirical risk. However, it encounters challenges in real-world applications due to its higher computational complexity. To mitigate the computational complexity of SVM, \citet{suykens1999least} introduced a variant called the least squares SVM (LSSVM). Unlike SVM, LSSVM uses a quadratic loss function rather than a hinge loss function, which enables the incorporation of equality constraints in the classification problem. Consequently, the solution can be achieved by solving a set of linear equations, eliminating the need to solve QPP. The computation time of LSSVM is significantly less compared to SVM, making it more efficient for large-scale problems. However, linear SVM and LSSVM only generate linear decision boundaries. If the data is not linearly separable, these models will perform poorly as they cannot accurately classify the data points. In contrast, kernel-based SVM and LSSVM have significantly impacted a wide range of application domains by effectively handling complex data structures \cite{shawe2004kernel, bishop2006pattern}. Kernel-based SVM and LSSVM models are firmly established with robust foundations in learning theory and optimization. Kernel functions allow SVM and LSSVM to operate in a high-dimensional feature space implicitly, enabling them to capture non-linear relationships between features. This flexibility makes SVM with kernels highly versatile and applicable to a wide range of data types and structures. Despite their robust mathematical foundations, kernel methods encounter difficulties when scaling to accommodate large datasets. Model selection in kernel methods, such as choosing the kernel function and its parameters, poses a non-trivial challenge. Furthermore, \citet{suykens2017deep} introduced restricted kernel machines (RKM) with the objective of closing the divide with neural network-based models and broadening the utility of kernel methods to tackle complex real-world problems. RKM leverages Legendre-Fenchel duality to acquire a representation of LSSVMs resembling the energy function of RBM \cite{hinton2006fast}. While RKM has been successfully applied in generative models \cite{pandey2021generative, pandey2022disentangled}, disentangled representations \cite{tonin2021unsupervised, tonin2021unsupervised} and classification \cite{houthuys2021tensor}, no previous study has investigated or utilized the use of RKM for anomaly detection (fault detection) and outlier detection.  

For fault and outlier detection using SVM and LSSVM, it's essential to gather datasets comprising both normal and various types of faulty data from the industrial process for training purposes. While normal operation data is readily available, acquiring faulty data sets poses challenges due to the following reasons: (1) inducing faults in the industrial process is costly; (2) the practical industry process involves numerous and diverse fault categories, making it impractical to include all possible faults for SVM training. However, if only data from one class is available for training, using SVM for related learning tasks becomes challenging. This scenario is known as novelty detection or one-class classification in the literature \cite{alam2020one, seliya2021literature}. \citet{scholkopf2001estimating} proposed the one-class SVM (OCSVM) as an alternative approach. The fundamental idea behind OCSVM is to generate a hyperplane that maximizes the margin between the origin and the samples of a specific class, primarily determined by the support vectors. Thus, OCSVM maintains the theoretical foundations of SVM and has become one of the most widely adopted techniques for one-class classification tasks. OCSVM has showcased efficacy across diverse domains, including network intrusion detection \cite{pang2022hybrid}, machine fault diagnosis \cite{zhao2020improved}, image annotation \cite{goh2005using}, music recommendation \cite{yepes2018listen} and so on. However, the standard OCSVM requires solving a QPP because of the inclusion of inequality constraints. This can lead to a less efficient optimization process. To overcome this issue, the least squares one-class SVM (LSOCSVM) \cite{choi2009least} is introduced, reforming the OCSVM optimization problem using an equality constraint and squared loss function. The solution to this restructured optimization problem can be obtained directly by solving a set of linear equations, bypassing the need for QPP, rendering it computationally more efficient than the standard OCSVM. Building on LSOCSVM, several powerful extensions have been proposed in recent years. Numerous variants of the OCSVM model are designed to enhance its generalization performance across various aspects, such as approximated LSOCSVM \cite{mygdalis2015large}, multi-view OCSVM method with privileged information learning (MOCPIL) \cite{xiao2024privileged}, multi-kernel correntropy based LSOCSVM (MKCLSOCSVM) \cite{zheng2023multikernel}, and so on. 

A primary drawback of LSOCSVM and many of its variants is their susceptibility to performance degradation when confronted with contaminated data, such as mislabeled observations. This challenge primarily arises due to the utilization of the squared loss function in these methods, which has been shown to be sensitive to outliers or noise that deviates significantly from a Gaussian distribution. To address this concern, researchers have turned to the correntropy loss as a replacement for the squared loss function in the design of LSOCSVM \cite{xing2020robust}. As correntropy \cite{chen2019effects, bako2018robustness} is a similarity measure that is insensitive to outliers, adopting it significantly enhances the robustness of LSOCSVM. Similarly, \citet{xing2022adaptive} proposed an adaptive loss function-based LSOCSVM. By appropriately tuning the shape parameter, this loss function can emulate various commonly employed robust loss functions \cite{barron2019general}, including Cauchy loss and $L_1$ loss, as special cases. Therefore, LSOCSVM employing this adaptive loss function provides increased flexibility and is anticipated to exhibit enhanced learning performance, particularly in scenarios involving one-class classification tasks with contaminated data. Motivated by the effectiveness of OCSVM and LSOCSVM in anomaly and fault and outlier detection, we propose one class RKM (OCRKM) within the framework of RKM. The OCRKM introduces a unique representation of kernel methods, integrating both visible and hidden variables, thereby establishing a connection between kernel methods and RBM \cite{hinton2006fast} through the reminiscent energy function. The OCRKM model is designed to detect and penalize outliers effectively. This unique representation enables OCRKM to achieve superior anomaly, fault and outlier detection capabilities compared to traditional OCSVMs. RKM projects data samples into a latent space and learns to distinguish normal data points from outliers by defining a boundary. To the best of our knowledge, this is the first time that the RKM framework has been employed in one-class classification to enhance their robustness. The following are the main highlights of this paper:
\begin{enumerate}
    \item In this paper, we propose a novel one-class restricted kernel machine (OCRKM) model. OCRKM eliminates the requirement for labeled two-class information during training, enabling it to effectively detect outliers and anomalies. 
     \item Thanks to the introduction of the Fenchel-Young inequality, we derive RKM formulation tailored for one-class classification. In RKM, data samples are projected into a latent space, where the model learns to establish a boundary that encloses normal data points while effectively excluding outliers.
    \item OCRKM utilizes an energy function similar to that of the RBM, incorporating both visible and hidden variables within a non-probabilistic framework.
    \item In OCRKM, the kernel trick is employed to map the data into a high-dimensional feature space. Within this space, the algorithm identifies a hyperplane that optimally separates the training instances using a regularized least squares approach.
    \item To evaluate the generalization performance of the proposed OCRKM model, we conduct experimental evaluation across $30$ UCI datasets. The empirical findings demonstrate that the proposed OCRKM model consistently outperforms numerous baseline models.
\end{enumerate}

The remaining structure of the paper is organized as follows. Section \ref{Related Work}, provides an overview of related work. We present the detailed mathematical formulation of the proposed OCRKM model in Section \ref{OCRKM}. Experimental results and analyses of proposed and existing models are discussed in Section \ref{Experiments and Results}. Finally, Section \ref{Conclusion and Future Work} gives the conclusions and potential future research directions.

\section{Related Work}
\label{Related Work}
This section offers a concise overview of the least square one-class support vector machine (LSOCSVM) and RKM, both of which are crucial in the development and formulation of our proposed OCRKM.
\subsection{LSOCSVM}
The main idea of LSOCSVM \cite{choi2009least} is to identify a hyperplane in the high-dimensional feature space that maximally separates the samples from the origin. Consider that the training set $S=\{x_i\}_{i=1}^N$ with \(x_i \in \mathbb{R}^{1 \times m}\), where \(m\) is the number of input samples. $\phi(\cdot)$ represents the feature mapping corresponding to the kernel function. The optimization problem of LSOCSVM is given as follows:
\begin{align}
\label{OCSVM:1}
    &\min\frac{1}{2}\|w\|^2 - \rho +  \frac{\mathcal{C}}{2}\sum_{i=1}^N \xi_i^2 \nonumber \\
    &\text{s.t.} \hspace{0.2cm} w\phi(x_i) = \rho - \xi_i, \hspace{0.2cm} i=1,2,\ldots, N,
\end{align}
where $\mathcal{C}$ is tunable parameter and $\xi_i$ are slack variables. The weight and bias of the separating hyperplane are denoted by $w$ and $\rho$, respectively. Then the Wolfe dual for \eqref{OCSVM:1} can be obtained as follows:
\begin{align}
\left[\begin{array}{c|c } 
	 K + I/\mathcal{C} & e  \\  
	\hline 
	 e^T    & 0  
\end{array}\right]
    \begin{bmatrix}
    \alpha \\ -\rho 
    \end{bmatrix}  =  \begin{bmatrix} 0 \\   1   \end{bmatrix}
\end{align}
where $\alpha$ is the Lagrangian multiplier and \(K(x_i, x_j) = \phi(x_i)^T\phi(x_j)\) denotes the kernel function.
\subsection{RKM}
Here, we provide an overview of the RKM classification model as described by \citet{suykens2017deep}, which is closely related to the well-known LSSVM \cite{suykens1999least} model. RKM employs the kernel trick to map the data into a high-dimensional feature space, where a linear separating hyperplane can be constructed. The objective function of RKM is as follows:
\begin{align}
\label{eq:1111}
      J = & \frac{\gamma}{2} Tr(W^TW) + \sum_{i=1}^N (1-(\phi(x_i)^TW+b)y_i)h_i - \frac{\eta}{2} \sum_{i=1}^Nh_i^2,
\end{align}
where $\gamma$ and $\eta$ are the regularization parameters, $b$ is the bias term, and $h$ is hidden features, respectively. The solution of equation \eqref{eq:1111} can be derived by taking partial derivatives of $J$ with respect to $W$, $b$, and $h_i$, and subsequently equating the resulting expressions to zero. For detailed derivation, refer to \citet{suykens2017deep}.
\section{Proposed One Class Restricted Kernel Machine (OCRKM)}
\label{OCRKM}
\begin{figure}[ht!]
    \centering
    \includegraphics[width=0.8\textwidth,height=6cm]{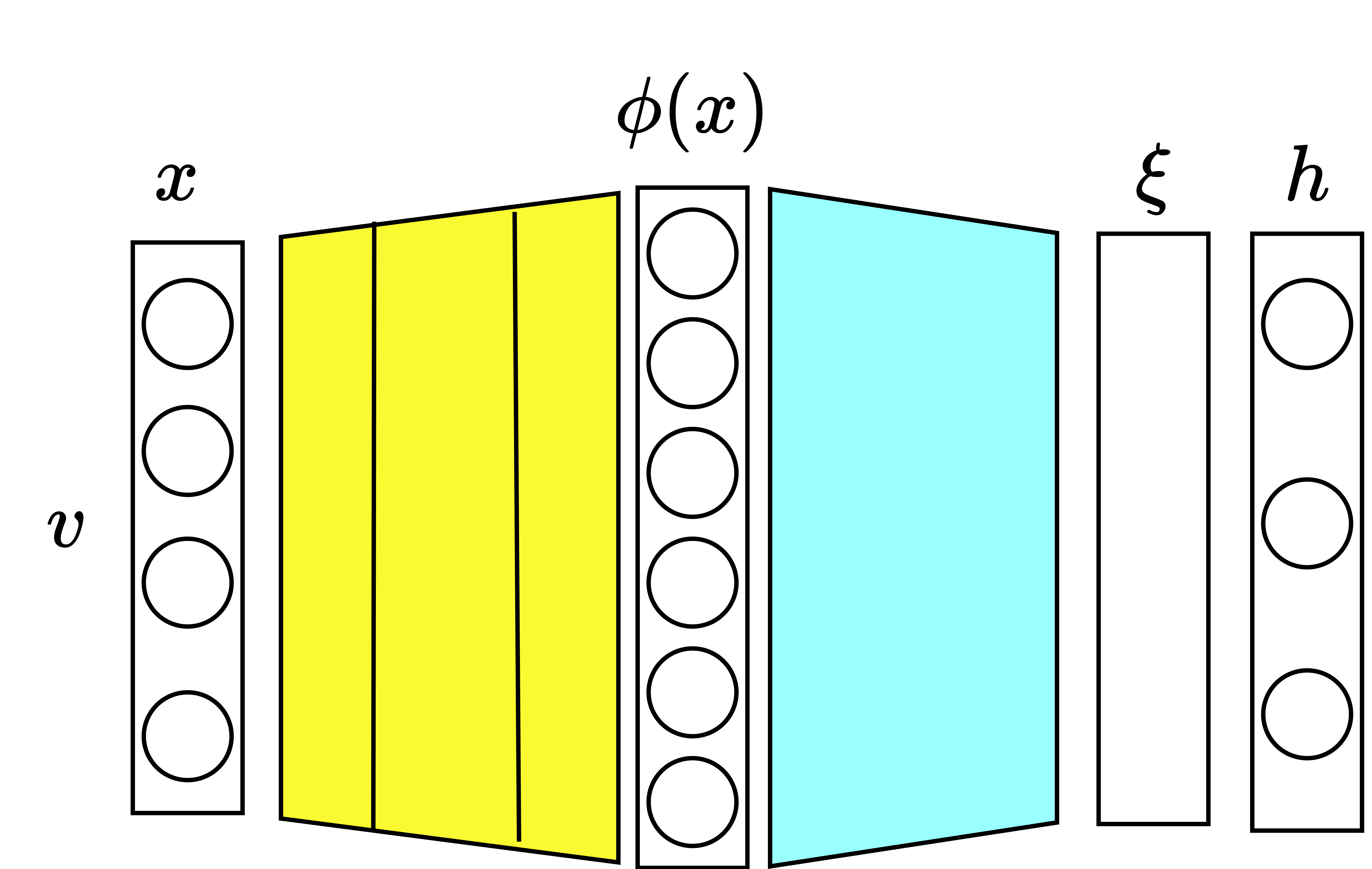}
    \caption{The geometrical structure of the One-Class Robust Kernel Machine (OCRKM) model involves mapping an input \( x \) from the original space to a higher-dimensional feature space using a non-linear mapping function \( \phi(x) \). In this feature space, the model learns projections \( \xi \) that capture the underlying structure of the data while being coupled with hidden features \( h \) in a latent space. These hidden features \( h \) help to model the intrinsic characteristics of the data distribution, enabling the model to distinguish between normal and abnormal data points. The coupling between the feature space projections \( \xi \) and the latent space features \( h \) allows the OCRKM model to effectively learn robust decision boundaries that are less sensitive to noise and outliers, improving its one-class classification performance.}
    \label{Geometrical structure of OCRKM model}
\end{figure}

This section offers an in-depth explanation of the proposed OCRKM model. Initially, we outline the general mathematical framework of the proposed OCRKM model, specifically designed to handle data from a single class. OCRKM offers a unique representation of kernel methods, incorporating both visible and hidden variables, akin to the energy function similar to RBM \cite{hinton2006fast}, thus linking kernel methods with RBM. OCRKM can be linked to the energy form expression of an RBM, with an interpretation in terms of hidden \( h \) and visible units \( v \). Similar to RBM, OCRKM is characterized by the scalar-valued energy function during both the training and prediction phases. In line with kernel methods, the optimality conditions of this objective function are achieved by solving a linear system or matrix decomposition during training. The geometrical representation of the proposed OCRKM is shown in Fig. \ref{Geometrical structure of OCRKM model}.

Assume that the function $\phi: x_i \rightarrow \phi(x_i)$ maps the training samples from the input space to a high-dimensional feature space.
The optimization problem of OCRKM  can be expressed as follows:

\begin{align}
\label{eq:1}
     \underset{V, \xi}{\min}~J(V, \xi) = & \underset{V, \xi}{\min}~\frac{\gamma}{2} Tr(V^TV)  - \rho +\frac{1}{2\eta}\sum_{i=1}^N \xi_i^T\xi_i \nonumber \\
   & \hspace{0.07cm} \textbf{s.t.}  \hspace{0.2cm} \xi_i = \rho - V^T \phi(x_i), \hspace{0.2cm}  \forall \hspace{0.1cm} i,
\end{align}
where $V$ is the interconnection matrix, $\xi$ is the error vector, $\rho$ denotes the bias term, $\gamma$, and $\eta$ are the tunable parameter, respectively. The link between LSOCSVM and RBM in OCRKM is established using the concept of the Legendre-Fenchel conjugate for quadratic functions \cite{rockafellar1974conjugate}. The OCRKM provides an upper bound of $J$ by introducing the hidden layer representations $h$ as follows \cite{rockafellar1974conjugate}:
\begin{align}
\label{eq:2}
    \frac{1}{2\eta}\xi^T\xi \geq \xi^Th - \frac{\eta}{2}h^Th, \hspace{0.2cm} \forall \hspace{0.1cm} \xi, h.
\end{align}
On employing \eqref{eq:2} in \eqref{eq:1}, we obtain:
\begin{align}
   \hspace{2cm} &J \geq \sum_{i=1}^N \xi_i^T h_i - \frac{\eta}{2} \sum_{i=1}^N h_i^Th_i + \frac{\gamma}{2} Tr(V^TV) - \rho \nonumber \\ 
   &  \textbf{s.t.} \hspace{0.2cm} \xi_i = \rho - V^T \phi(x_i), \hspace{0.2cm}  \forall \hspace{0.1cm} i.
   \end{align}
Substitute the constraints to obtain the following tight upper-bound:
\begin{align}
 \implies   &  J \geq \sum_{i=1}^N (\rho - \phi(x_i)^TV)h_i - \frac{\eta}{2}\sum_{i=1}^N h_i^Th_i + \frac{\gamma}{2} Tr(V^TV) - \rho  = \hat{J}. 
\end{align}
Next, we proceed by examining the stationary points of $\hat{J}$. 
\begin{align}
    & \frac{\partial \hat{J}}{\partial V} = 0 \implies V = \frac{1}{\gamma}\sum_{i=1}^N \phi(x_i) h_i^T, \\
    & \frac{\partial \hat{J}}{\partial h_i} = 0 \implies \rho = V^T \phi(x_i) + \eta h_i, \hspace{0.2cm}  \forall \hspace{0.1cm} i \\
    & \frac{\partial \hat{J}}{\partial \rho} = 0 \implies \sum_{i=1}^N h_i = 1.
\end{align}
Eliminating the weight $V$, we obtain the following linear system:
\begin{align}
\left[\begin{array}{c|c } 
	\frac{1}{\gamma} \sum_{i=1}^N \sum_{j=1}^N \phi(x_i) \phi(x_j)^T + \eta I_N & -1_N  \\  
	\hline 
	 1_N^T    & 0  
\end{array}\right]
    \begin{bmatrix}
    H \\ \rho 
    \end{bmatrix}  =  \begin{bmatrix} 0 \\   1_N   \end{bmatrix},
\end{align}
where $H = [h_1, h_2, \ldots, h_N]$, $I_N$ represents the matrix of ones of suitable dimensions, and $1_N$ represents the vector whose entries are all 1, with appropriate dimensions.

We also use the notation \(K(x_i, x_j)\) to represent the \(i\)-th row and \(j\)-th column entry of the kernel matrix \(K\), and is defined as: \(K(x_i, x_j) = \phi(x_i)^T\phi(x_j)\). 
\begin{align}
\label{eq:10}
\left[\begin{array}{c|c } 
	\frac{1}{\gamma} K + \eta I_N & -1_N  \\  
	\hline 
	 1_N^T    & 0  
\end{array}\right]
    \begin{bmatrix}
    H \\ \rho 
    \end{bmatrix}  =  \begin{bmatrix} 0 \\   1_N   \end{bmatrix}.
\end{align}
Finally, the decision function with dual representation is given as follows:
\begin{align}
\label{eq:11}
    f(x) = \frac{1}{\gamma} \sum_{i}h_iK(x_i, x) - \rho.
\end{align}
Then, the class label for the test instance $x$ is determined by
\begin{align}
\label{eq:12}
    \hat{y} = sign(f(x)),
\end{align}
where $sign(\cdot)$ denotes the sign function. The algorithm of the proposed OCRKM model is described in Algorithm \ref{OCRKM training and prediction}.

\begin{algorithm}
\caption{Training and prediction of the proposed OCRKM.}
\label{OCRKM training and prediction}
\textbf{Input:} Let $\{x_i\}_{i=1}^N$ be the input training dataset and the trade-off parameters $\gamma$ and $\eta$, respectively.
\begin{algorithmic}[1]
\STATE Find the kernel matrix using $K=\sum_{i=1}^N \phi(x_i)^T \phi(x_i)$.\\
\STATE Calculate $H$ and $\rho$ using Eq. \eqref{eq:10}.\\
\STATE Find the decision function with dual representation using Eq. \eqref{eq:11}.
\STATE Testing sample $x$ is classified into class $+1$ or $-1$ using Eq. \eqref{eq:12}.
\end{algorithmic}
\textbf{Output:} $\hat{y}(x).$
\end{algorithm}
\section{Experimental Results}
\label{Experiments and Results}
To evaluate the effectiveness of the proposed OCRKM model, we conduct a comparative analysis along with baseline models using benchmark datasets from the UCI \cite{dua2017uci} repository.
\subsection{Experimental Setup}
The experimental setup includes a PC with an Intel(R) Xeon(R) Gold 6226R CPU running at $2.90$ GHz, $128$ GB of RAM, and operating on the Windows $11$ platform. Python $3.11$ is used for the execution of all models. The dataset is randomly splitted into training and testing subsets at a ratio of $70:30$, respectively. We utilize $5$-fold cross-validation along with a grid search approach to fine-tune the hyperparameters of the models within the ranges: $\eta = \gamma = \{10^{-5}, 10^{-4}, \ldots, 10^{5}\}$. We employ the Gaussian kernel for non-linear cases, which is defined as $\mathscr{K}(x_i, x_j) = e^{\frac{-1}{2\sigma^2}\|x_i - x_i\|^2}$. The Gaussian kernel parameter $\sigma$ is chosen from within the range $\{2^{-5}, 2^{-4}, \ldots, 2^{5}\}$. For OCSVM \cite{scholkopf2001estimating} and LSOCSVM \cite{choi2009least}, the hyperparameter $C$ is selected from $\{10^{-5}, 10^{-4}, \ldots, 10^{5}\}$. For MKCLSOCSVM \cite{zheng2023multikernel}, the hyperparameter $C_1$, $C_2$, $\sigma_1$, $\sigma_2$, $\gamma$, and $\lambda$ are selected from $\{10^{-5}, 10^{-4}, \ldots, 10^{5}\}$. For robust OCSVM (ROCSVM) \cite{xing2020robust}, the hyperparameter $C$ is selected similar to OCSVM and the hyperparameter $\eta$ is selected from $\{0.001, 0.01, 0.1, 0.2, . . ., 1, 1.2, 1.5, 2, 4, 7, 10\}$. For OCELM \cite{leng2015one}, the hyperparameter $C$ is selected similar to OCSVM and the number of hidden nodes is chosen from the range $3:20:203$.

\subsection{Experiments on Real World UCI Datasets}
In this subsection, we conduct a comparative analysis between the proposed OCRKM model and the baseline OCSVM \cite{scholkopf2001estimating}, LSOCSVM \cite{choi2009least}, MKCLSOCSVM \cite{zheng2023multikernel}, robust OCSVM (ROCSVM) \cite{xing2020robust}, and OCELM \cite{leng2015one} models using $30$ UCI benchmark datasets, which encompass diverse domains and sample sizes. The comparison is conducted using a range of statistical metrics and tests, including accuracy, ranking, the Friedman test, and the Nemenyi post hoc test.  

\begin{table}[ht!]
\centering
    \caption{Performance comparison of the proposed OCRKM along with the baseline models over UCI datasets.}
    \label{Average ACC and average rank for proteins datasets}
    \resizebox{1.0\linewidth}{!}{
\begin{tabular}{lcccccc}
\hline
Dataset & OCSVM \cite{scholkopf2001estimating} & LSOCSVM \cite{choi2009least} & MKCLSOCSVM \cite{zheng2023multikernel} & ROCSVM \cite{xing2020robust} & OCELM \cite{leng2015one} & OCRKM \\
 & $(ACC, Std)$ & $(ACC, Std)$ & $(ACC, Std)$ & $(ACC, Std)$ & $(ACC, Std)$ & $(ACC, Std)$ \\ \hline
acute\_inflammation & $(74.04, 0.28)$ & $(67.76, 4.87)$ & $(60.41, 0.05)$ & $(62.13, 7.12)$ & $(85.96, 12.37)$ & $(\textbf{86.15}, 3.81)$ \\
acute\_nephritis & $(83.18, 2.34)$ & $(79.23, 4.17)$ & $(33.85, 0.04)$ & $(62.73, 5.23)$ & $(82.73, 13.5)$ & $(\textbf{85.71}, 4.71)$ \\
blood & $(60.95, 2.84)$ & $(64.55, 0.16)$ & $(51.01, 0)$ & $(46.43, 1.42)$ & $(66.3, 1.98)$ & $(\textbf{70.68}, 0.37)$ \\
breast\_cancer & $(56.48, 8.04)$ & $(71.83, 2.48)$ & $(52.17, 0.01)$ & $(55.82, 0.92)$ & $(73.04, 5.77)$ & $(\textbf{73.36}, 3.27)$ \\
breast\_cancer\_wisc & $(76.37, 4.26)$ & $(55.17, 1.2)$ & $(71.58, 0.03)$ & $(64.3, 1.45)$ & $(71.27, 23.64)$ & $(\textbf{79.92}, 1.67)$ \\
breast\_cancer\_wisc\_diag & $(64.95, 1.72)$ & $(65.4, 3.09)$ & $(67.81, 0.06)$ & $(67.17, 2.02)$ & $(43.75, 5.49)$ & $(\textbf{73.7}, 2.44)$ \\
breast\_cancer\_wisc\_prog & $(46.21, 4.88)$ & $(64.85, 0.28)$ & $(45.2, 0)$ & $(48.62, 1.89)$ & $(84, 11.38)$ & $(\textbf{84.42}, 0.58)$ \\
conn\_bench\_sonar\_mines\_rocks & $(\textbf{73.5}, 4.27)$ & $(41.3, 4.07)$ & $(56.74, 0.03)$ & $(72, 4.18)$ & $(70.75, 7.16)$ & $(73.03, 0.82)$ \\
credit\_approval & $(71.65, 2.83)$ & $(70.5, 1.9)$ & $(60.61, 0.06)$ & $(77.73, 2.04)$ & $(75.91, 10.21)$ & $(\textbf{78.88}, 1.63)$ \\
echocardiogram & $(61.36, 8.84)$ & $(76.04, 5.66)$ & $(65.48, 0.01)$ & $(61.82, 2.96)$ & $(72.73, 13.02)$ & $(\textbf{79.19}, 1.4)$ \\
fertility & $(33.6, 2.3)$ & $(61.82, 2.49)$ & $(62.91, 0.02)$ & $(35.2, 7.69)$ & $(\textbf{98.4}, 3.58)$ & $(89.52, 1.61)$ \\
heart\_hungarian & $(63.14, 7.92)$ & $(52.55, 0.58)$ & $(63.58, 0.04)$ & $(63.14, 0.54)$ & $(68.96, 18.59)$ & $(\textbf{69.81}, 4.84)$ \\
hepatitis & $(81.5, 1.23)$ & $(65.29, 1.49)$ & $(66.82, 0.01)$ & $(\textbf{87}, 7.1)$ & $(80.25, 7.78)$ & $(83.77, 0.88)$ \\
horse\_colic & $(54.1, 3.56)$ & $(63.13, 1.69)$ & $(69.4, 0.01)$ & $(58.59, 6)$ & $(75.9, 12.7)$ & $(\textbf{75.98}, 5.59)$ \\
ilpd\_indian\_liver & $(80.5, 3.48)$ & $(54.1, 0.86)$ & $(56.96, 0.01)$ & $(54.21, 0.6)$ & $(77.1, 1.04)$ & $(\textbf{80.73}, 1.02)$ \\
ionosphere & $(65.3, 2.49)$ & $(64.25, 4.41)$ & $(79.84, 0.03)$ & $(82.13, 0.95)$ & $(\textbf{85.79}, 3.38)$ & $(84.59, 3.38)$ \\
molec\_biol\_promoter & $(73.27, 8.97)$ & $(73.95, 3.03)$ & $(53.49, 0.05)$ & $(75.81, 2.08)$ & $(77.91, 5.3)$ & $(\textbf{79.29}, 3.03)$ \\
monks\_2 & $(62.25, 1.78)$ & $(73.45, 0.83)$ & $(56.19, 0.02)$ & $(62.37, 0.82)$ & $(80.56, 1.25)$ & $(\textbf{81.01}, 2.03)$ \\
monks\_3 & $(85.9, 5.25)$ & $(64.05, 8.89)$ & $(63.95, 0.01)$ & $(76.02, 1.01)$ & $(\textbf{78.81}, 0.83)$ & $(75.17, 1.55)$ \\
musk\_1 & $(80.37, 1.77)$ & $(80.06, 0.82)$ & $(71.19, 0.87)$ & $(65.06, 6.76)$ & $(80.42, 18.09)$ & $(\textbf{80.49}, 0.19)$ \\
oocytes\_merluccius\_nucleus\_4d & $(90.61, 2.47)$ & $(80.81, 0.29)$ & $(61.83, 0.73)$ & $(85.36, 0.44)$ & $(90.44, 2.34)$ & $(\textbf{92.64}, 0.31)$ \\
oocytes\_trisopterus\_nucleus\_2f & $(63.91, 4.17)$ & $(84.08, 6.56)$ & $(58.99, 0.75)$ & $(67.44, 1.81)$ & $(75.81, 0.84)$ & $(\textbf{76.12}, 1.33)$ \\
pittsburg\_bridges\_T\_OR\_D & $(55.09, 5.89)$ & $(57.69, 5.02)$ & $(47.5, 0.59)$ & $(63.85, 3.22)$ & $(71.43, 7.72)$ & $(\textbf{73.7}, 5.02)$ \\
planning & $(69.49, 3.16)$ & $(54.04, 2.29)$ & $(55.46, 0.81)$ & $(54.74, 0.79)$ & $(\textbf{75.46}, 1.61)$ & $(72.75, 0.96)$ \\
spectf & $(86.22, 1.5)$ & $(89.42, 1.21)$ & $(80.27, 0.91)$ & $(\textbf{93.33}, 0.61)$ & $(87.68, 11.37)$ & $(91.7, 1.32)$ \\
statlog\_heart & $(80.23, 4.24)$ & $(70.98, 0.88)$ & $(52.98, 0.88)$ & $(70.2, 0.54)$ & $(69.61, 8.85)$ & $(\textbf{81.62}, 2.97)$ \\
tic\_tac\_toe & $(60.72, 1.46)$ & $(68.46, 0.65)$ & $(65.99, 0.81)$ & $(70.78, 1.25)$ & $(\textbf{77.41}, 4.68)$ & $(71.78, 1.68)$ \\
titanic & $(70, 1.9)$ & $(58.7, 2.83)$ & $(\textbf{78.71}, 0.71)$ & $(62.23, 1.29)$ & $(74.76, 10.88)$ & $(75.87, 0.99)$ \\
trains & $(57.06, 3.11)$ & $(72.9, 4.71)$ & $(60, 0.34)$ & $(61.45, 12.37)$ & $(\textbf{80}, 32.6)$ & $(73.33, 11.18)$ \\
vertebral\_column\_2clases & $(\textbf{85.89}, 4.94)$ & $(56.47, 1.78)$ & $(60.55, 0.65)$ & $(56.86, 2.08)$ & $(84.11, 8.44)$ & $(77.91, 1.28)$ \\ \hline
Average ACC & $68.93$ & $66.76$ & $61.05$ & $65.48$ & $77.24$ & $\textbf{79.09}$ \\ \hline
Average Std & $3.73$ & $2.64$ & $0.28$ & $2.91$ & $8.88$ & $\textbf{2.4}$ \\ \hline
Average Rank & $3.85$ & $4.30$ & $4.77$ & $4.02$ & $2.57$ & $\textbf{1.50}$ \\ \hline
\end{tabular}}
\end{table}

The experimental results, including accuracy ($ACC$), standard deviation ($Std$), and ranks, are presented in Table \ref{Average ACC and average rank for proteins datasets}. The accuracy-based comparison demonstrates that our proposed OCRKM model surpasses the baseline OCSVM, LSOCSVM, MKCLSOCSVM, ROCSVM, and OCELM models on the majority of datasets. From Table \ref{Average ACC and average rank for proteins datasets}, we observe that our proposed OCRKM model demonstrates the highest average ACC, achieving $79.09\%$. In contrast, the average ACC of the OCSVM, LSOCSVM, MKCLSOCSVM, ROCSVM, and OCELM models are $68.93\%$, $66.76\%$, $61.05\%$, $65.48\%$, and $77.24\%$ respectively. This comparison shows the superior and robust performance of our proposed OCRKM models. Sometimes, a model's average accuracy metric may be skewed by exceptional performance in one dataset that compensates for shortcomings across multiple datasets, it could potentially introduce bias into the measure. As a result, we utilize a ranking methodology to evaluate the efficacy of the compared models. In this method, every classifier is given a rank, with the model showing the best performance getting a lower rank, while the one with the poorest performance gets a higher rank. To evaluate $k$ models across $M$ datasets, we define $r_j^i$ as the rank of the $j^{th}$ model on the $i^{th}$ dataset. The average rank of the $j^{th}$ models is given by $\mathcal{R}_j = \frac{1}{M}\sum_{i=1}^Mr_j^i$. From the last row of Table \ref{Average ACC and average rank for proteins datasets}, we note the average ranks of the proposed OCRKM model, as well as the baseline OCSVM, LSOCSVM, MKCLSOCSVM, ROCSVM, and OCELM models are $3.85$, $4.30$, $4.77$, $4.02$, $2.57$, and $1.50$ respectively. The proposed OCRKM model achieves the lowest average rank, indicating superior performance compared to all other models, thus demonstrating its superior generalization capability. Now, we conduct the Friedman test \cite{demvsar2006statistical} to ascertain whether there are statistically significant differences among the models. Under the null hypothesis, it is posited that all models possess an identical average rank, signifying an equivalent level of performance. The Friedman statistic follows the chi-squared distribution (\( \chi^2_F \)) with (\( k - 1 \)) degrees of freedom (d.o.f), and its computation involves: $\chi^2_F = \frac{12M}{k(k+1)} \left[\sum_j \mathcal{R}_j^2 -\frac{k(k+1)^2}{4} \right]$. The $F_F$ statistic is calculated as $F_F = \frac{(M-1)\chi_F^2}{M(k-1) - \chi_F^2}$,  where the $F$-distribution has degrees of freedom (\( k - 1 \)) and \( (M - 1) \times (k - 1) \). In our case, we have $p=6$ and $M=30$, thus the calculated values are $\chi_F^2 = 64.98$ and $F_F = 22.16$. At a $5\%$ level of significance, the critical value $F_{F} (5, 145) = 2.276$. We reject the null hypothesis as $22.16 > 2.276$. Hence, there exists a statistically significant difference among the compared models. Next, we employ the Nemenyi post hoc test to explore the pairwise variances between the models. The critical difference ($C.D.$) value is calculated as $C.D = q_\alpha \sqrt{\frac{k(k+1)}{6M}}$. With \( q_{\alpha} = 2.850 \) for $6$ models at a $5\%$ significance level, the resulting \( C.D. \) is \( 1.377 \). The average rank differences between the proposed OCRKM model and the baseline OCSVM, LSOCSVM, MKCLSOCSVM, ROCSVM, and OCELM models are $2.35$, $2.80$, $3.27$, $2.52$, and $1.07$, respectively. The Nemenyi post hoc test confirms that the proposed OCRKM model is significantly superior to the baseline OCSVM, LSOCSVM, MKCLSOCSVM, and ROCSVM models. However, the average rank of the proposed OCRKM surpasses the OCELM. Hence, the proposed OCRKM model showcases superior performance against the baseline models.
\begin{table}[ht!]
    \centering
    \caption{Pairwise win-tie-loss test of proposed OCRKM models along with baseline models on UCI datasets.}
   \label{Pairwise Win-tie using linear kernel}
    \resizebox{1\textwidth}{!}{
\begin{tabular}{lccccc}
\hline
 & OCSVM \cite{scholkopf2001estimating} & LSOCSVM  \cite{choi2009least} & MKCLSOCSVM \cite{zheng2023multikernel} & ROCSVM \cite{xing2020robust} & OCELM \cite{leng2015one} \\ \hline
LSOCSVM \cite{choi2009least} & {[}$14, 0, 16${]} &  &  &  &  \\ 
MKCLSOCSVM \cite{zheng2023multikernel} & {[}$9, 0, 21${]} & {[}$12, 0, 18${]} &  &  &  \\
ROCSVM \cite{xing2020robust} & {[}$15, 1, 14${]} & {[}$17, 0, 13${]} & {[}$18, 0, 12${]} &  &  \\
OCELM \cite{leng2015one} & {[}$20, 0, 10${]} & {[}$25, 0, 5${]} & {[}$27, 0, 3${]} & {[}$24, 0, 6${]} &  \\
OCRKM & {[}$27, 0, 3${]} & {[}$29, 0, 1${]} & {[}$29, 0, 1${]} & {[}$27, 0, 3${]} & {[}$23, 0, 7${]} \\ \hline
\end{tabular}}
\end{table}

Furthermore, we employ the pairwise win-tie-loss (W-T-L) sign test. This test operates on the assumption, under the null hypothesis, that two models perform equally and are anticipated to win in $M/2$ datasets, where $M$ denotes the dataset count. If a model wins on roughly \( \frac{M}{2} + 1.96 \frac{\sqrt{M}}{2} \) datasets, it is considered significantly superior. If there is an even count of ties between the two models, these ties are evenly distributed between them. However, if the number of ties is odd, one tie is ignored, and the remaining ties are divided among the specified classifiers. In this scenario, with \( M = 30 \), if one of the models attains wins of at least \( 20.36 \), then a significant difference between the models exists. Table \ref{Pairwise Win-tie using linear kernel} presents a comparative analysis of the proposed OCRKM model alongside the baseline models. It delineates their performance regarding pairwise wins, ties, and losses across UCI datasets. In Table \ref{Pairwise Win-tie using linear kernel}, the entry $[x, y, z]$ indicates the frequency with which the model listed in the row wins $x$ times, ties $y$ times, and loses $z$ times when compared to the model listed in the corresponding column. In Table \ref{Pairwise Win-tie using linear kernel}, the W-T-L outcomes of the models are evaluated pairwise. The proposed OCRKM model have achieved $27$ wins (against OCSVM), $29$ wins (against LSOCSVM), $29$ wins (against MKCLSOCSVM), and $27$ wins (against ROCSVM) out of $30$ datasets. The proposed OCRKM model achieves a statistically significant difference from the baseline models.
\begin{figure*}[ht!]
\begin{minipage}{.50\linewidth}
\centering
\subfloat[acute\_inflammation]{\label{fig1}\includegraphics[scale=0.33]{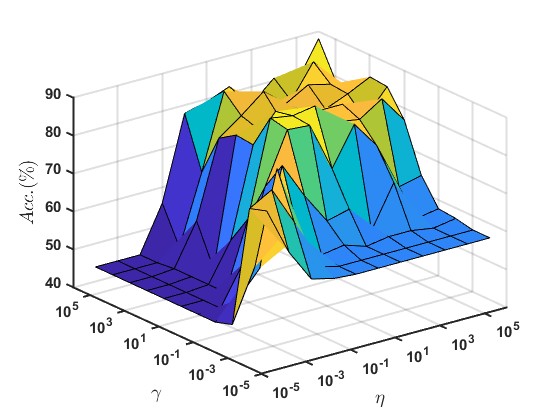}}
\end{minipage}
\begin{minipage}{.50\linewidth}
\centering
\subfloat[ilpd\_indian\_liver]{\label{fig2}\includegraphics[scale=0.33]{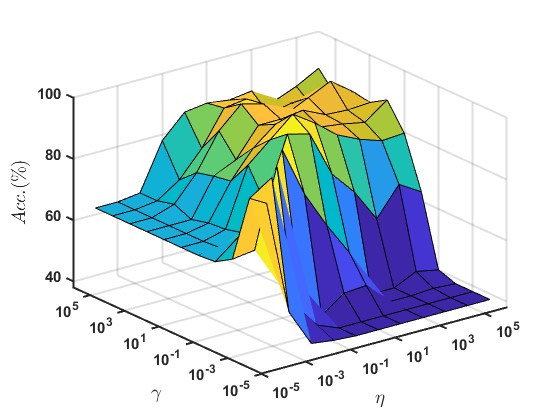}}
\end{minipage}
\par\medskip
\begin{minipage}{.50\linewidth}
\centering
\subfloat[molec\_biol\_promoter]{\label{fig3}\includegraphics[scale=0.33]{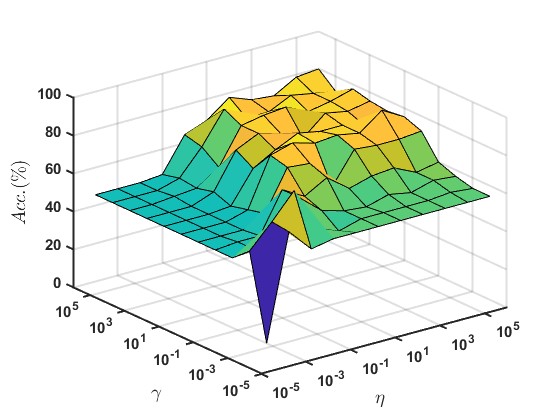}}
\end{minipage}
\begin{minipage}{.50\linewidth}
\centering
\subfloat[vertebral\_column\_2clases]{\label{fig4}\includegraphics[scale=0.33]{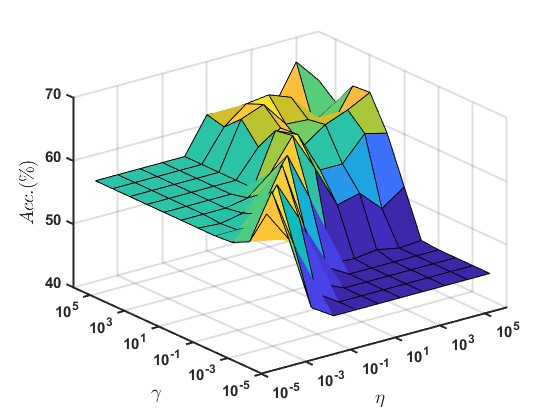}}
\end{minipage}
\caption{Effect of hyperparameters $\eta$ and $\gamma$ on the performance of the proposed OCRKM model.}
\label{effect of sigma parameter}
\end{figure*}
\subsection{Sensitivity Analyses}
To thoroughly understand the robustness of the proposed OCRKM model, it is essential to analyze their sensitivity to hyperparameters $\eta$ and $\gamma$. This thorough exploration enables us to identify the configuration that maximizes predictive accuracy and enhances the model's resilience when confronted with unseen data. Figure \ref{effect of sigma parameter} illustrates a significant fluctuation in the model's accuracy across various $\eta$ and $\gamma$ values, underscoring the sensitivity of our model's performance to these specific hyperparameters. From Figs. \ref{fig1} and \ref{fig2}, the optimal performance of the proposed OCRKM model is observed within the $\eta$ and $\gamma$ ranges of $10^{-1}$ to $10^{5}$ and $10^{1}$ to $10^{5}$, respectively. From Figs. \ref{fig3} and \ref{fig4}, the accuracy of the proposed OCRKM model archives the maximum when $\eta$ and $\gamma$ ranges of $10^{1}$ to $10^{5}$, respectively. Therefore, we recommend using $\eta$ and $\gamma$ from the range $10^{1}$ to $10^{5}$ for efficient results, although fine-tuning may be necessary depending on the dataset's characteristics for the proposed OCRKM model to achieve optimal generalization performance. 
\section{Conclusions and Future Directions}
\label{Conclusion and Future Work}
In this paper, we proposed a novel one-class restricted kernel machine (OCRKM) model. OCRKM adopts an energy function akin to that used in the RBM. This energy function encompasses both hidden and latent variables within the framework of a non-probabilistic setting. In OCRKM, the kernel trick is utilized to project the data into a high-dimensional feature space. Within this space, the algorithm identifies a hyperplane that optimally separates the training instances through a regularized least squares approach, establishing a boundary that encloses normal data points while effectively excluding outliers for one class classification. Our experimentation entailed testing the proposed OCRKM model on benchmark datasets from UCI repositories, comparing them against $5$ state-of-the-art models. Our experimental results underscore the superior performance of the proposed OCRKM model, positioning it as the top-performing model. The proposed model exhibits outstanding performance by achieving an accuracy improvement of up to $1.85\%$ compared to the baseline models. Furthermore, statistical analyses including ranking, the Friedman test, the Nemenyi post hoc test, and the win-tie-loss sign test confirm the significantly enhanced robustness of our proposed model in comparison to the baseline models. Our proposed models have shown exceptional performance in single-view datasets. However, their effectiveness in multiview problems has yet to be evaluated. Future research should focus on adapting these models for multiview problems. Moreover, an essential research direction involves extending the models to accommodate datasets with multiview while simultaneously reducing computational complexity.  
\section*{Acknowledgement}
This study receives support from the Science and Engineering Research Board (SERB) through the Mathematical Research Impact-Centric Support (MATRICS) scheme Grant No. MTR/2021/000787. M. Sajid acknowledges the Council of Scientific and Industrial Research (CSIR), New Delhi, for providing fellowship under grants 09/1022(13847)/2022-EMR-I. The authors gratefully acknowledge the invaluable support provided by the Indian Institute of Technology Indore.
\bibliography{refs.bib}

\begin{thebibliography}{34}
\providecommand{\natexlab}[1]{#1}
\providecommand{\url}[1]{\texttt{#1}}
\expandafter\ifx\csname urlstyle\endcsname\relax
  \providecommand{\doi}[1]{doi: #1}\else
  \providecommand{\doi}{doi: \begingroup \urlstyle{rm}\Url}\fi

\bibitem[Alam et~al.(2020)Alam, Sonbhadra, Agarwal, and Nagabhushan]{alam2020one}
S.~Alam, S.~K. Sonbhadra, S.~Agarwal, and P.~Nagabhushan.
\newblock One-class support vector classifiers: A survey.
\newblock \emph{Knowledge-Based Systems}, 196:\penalty0 105754, 2020.

\bibitem[Bako(2018)]{bako2018robustness}
L.~Bako.
\newblock Robustness analysis of a maximum correntropy framework for linear regression.
\newblock \emph{Automatica}, 87:\penalty0 218--225, 2018.

\bibitem[Barron(2019)]{barron2019general}
J.~T. Barron.
\newblock A general and adaptive robust loss function.
\newblock In \emph{Proceedings of the IEEE/CVF Conference on Computer Vision and Pattern Recognition}, pages 4331--4339, 2019.

\bibitem[Bishop and Nasrabadi(2006)]{bishop2006pattern}
C.~M. Bishop and N.~M. Nasrabadi.
\newblock \emph{Pattern Recognition and Machine Learning}, volume~4.
\newblock Springer, 2006.

\bibitem[Chen et~al.(2019)Chen, Xing, Zhao, Du, and Principe]{chen2019effects}
B.~Chen, L.~Xing, H.~Zhao, S.~Du, and J.~C. Principe.
\newblock Effects of outliers on the maximum correntropy estimation: A robustness analysis.
\newblock \emph{IEEE Transactions on Systems, Man, and Cybernetics: Systems}, 51\penalty0 (6):\penalty0 4007--4012, 2019.

\bibitem[Choi(2009)]{choi2009least}
Y.-S. Choi.
\newblock Least squares one-class support vector machine.
\newblock \emph{Pattern Recognition Letters}, 30\penalty0 (13):\penalty0 1236--1240, 2009.

\bibitem[Cortes and Vapnik(1995)]{cortes1995support}
C.~Cortes and V.~Vapnik.
\newblock Support-vector networks.
\newblock \emph{Machine Learning}, 20:\penalty0 273--297, 1995.

\bibitem[Dem{\v{s}}ar(2006)]{demvsar2006statistical}
J.~Dem{\v{s}}ar.
\newblock Statistical comparisons of classifiers over multiple data sets.
\newblock \emph{The Journal of Machine Learning Research}, 7:\penalty0 1--30, 2006.

\bibitem[Dua and Graff(2017)]{dua2017uci}
D.~Dua and C.~Graff.
\newblock U{CI} machine learning repository.
\newblock \emph{Available: http://archive.ics.uci.edu/ml}, 2017.

\bibitem[Goh et~al.(2005)Goh, Chang, and Li]{goh2005using}
K.-S. Goh, E.~Y. Chang, and B.~Li.
\newblock Using one-class and two-class svms for multiclass image annotation.
\newblock \emph{IEEE Transactions on Knowledge and Data Engineering}, 17\penalty0 (10):\penalty0 1333--1346, 2005.

\bibitem[Hinton et~al.(2006)Hinton, Osindero, and Teh]{hinton2006fast}
G.~E. Hinton, S.~Osindero, and Y.-W. Teh.
\newblock A fast learning algorithm for deep belief nets.
\newblock \emph{Neural Computation}, 18\penalty0 (7):\penalty0 1527--1554, 2006.

\bibitem[Houthuys and Suykens(2021)]{houthuys2021tensor}
L.~Houthuys and J.~A. Suykens.
\newblock Tensor-based restricted kernel machines for multi-view classification.
\newblock \emph{Information Fusion}, 68:\penalty0 54--66, 2021.

\bibitem[Leng et~al.(2015)Leng, Qi, Miao, Zhu, and Su]{leng2015one}
Q.~Leng, H.~Qi, J.~Miao, W.~Zhu, and G.~Su.
\newblock One-class classification with extreme learning machine.
\newblock \emph{Mathematical Problems in Engineering}, 2015, 2015.

\bibitem[Li and Guan(2008)]{li2008joint}
Y.~Li and C.~Guan.
\newblock Joint feature re-extraction and classification using an iterative semi-supervised support vector machine algorithm.
\newblock \emph{Machine Learning}, 71:\penalty0 33--53, 2008.

\bibitem[Mygdalis et~al.(2015)Mygdalis, Alexandros, Tefas, and Pitas]{mygdalis2015large}
V.~Mygdalis, I.~Alexandros, A.~Tefas, and I.~Pitas.
\newblock Large-scale classification by an approximate least squares one-class support vector machine ensemble.
\newblock In \emph{2015 IEEE Trustcom/BigDataSE/ISPA}, volume~2, pages 6--10. IEEE, 2015.

\bibitem[Pal and Mather(2005)]{pal2005support}
M.~Pal and P.~M. Mather.
\newblock Support vector machines for classification in remote sensing.
\newblock \emph{International Journal of Remote Sensing}, 26\penalty0 (5):\penalty0 1007--1011, 2005.

\bibitem[Pandey et~al.(2021)Pandey, Schreurs, and Suykens]{pandey2021generative}
A.~Pandey, J.~Schreurs, and J.~A. Suykens.
\newblock Generative restricted kernel machines: A framework for multi-view generation and disentangled feature learning.
\newblock \emph{Neural Networks}, 135:\penalty0 177--191, 2021.

\bibitem[Pandey et~al.(2022)Pandey, Fanuel, Schreurs, and Suykens]{pandey2022disentangled}
A.~Pandey, M.~Fanuel, J.~Schreurs, and J.~A. Suykens.
\newblock Disentangled representation learning and generation with manifold optimization.
\newblock \emph{Neural Computation}, 34\penalty0 (10):\penalty0 2009--2036, 2022.

\bibitem[Pang et~al.(2022)Pang, Pu, and Li]{pang2022hybrid}
J.~Pang, X.~Pu, and C.~Li.
\newblock A hybrid algorithm incorporating vector quantization and one-class support vector machine for industrial anomaly detection.
\newblock \emph{IEEE Transactions on Industrial Informatics}, 18\penalty0 (12):\penalty0 8786--8796, 2022.

\bibitem[Richhariya and Tanveer(2018)]{richhariya2018eeg}
B.~Richhariya and M.~Tanveer.
\newblock E{EG} signal classification using universum support vector machine.
\newblock \emph{Expert Systems with Applications}, 106:\penalty0 169--182, 2018.

\bibitem[Rockafellar(1974)]{rockafellar1974conjugate}
R.~T. Rockafellar.
\newblock \emph{Conjugate Duality and Optimization}.
\newblock SIAM, 1974.

\bibitem[Sajid et~al.(2024, https://doi.org/10.1002/widm.1546)Sajid, Sharma, Beheshti, and Tanveer]{Sajid2024smc}
M.~Sajid, R.~Sharma, I.~Beheshti, and M.~Tanveer.
\newblock Decoding cognitive health using machine learning: A comprehensive evaluation for diagnosis of significant memory concern.
\newblock \emph{WIREs Data Mining and Knowledge Discovery}, page e1546, 2024, https://doi.org/10.1002/widm.1546.
\newblock \doi{https://doi.org/10.1002/widm.1546}.

\bibitem[Sch{\"o}lkopf et~al.(2001)Sch{\"o}lkopf, Platt, Shawe-Taylor, Smola, and Williamson]{scholkopf2001estimating}
B.~Sch{\"o}lkopf, J.~C. Platt, J.~Shawe-Taylor, A.~J. Smola, and R.~C. Williamson.
\newblock Estimating the support of a high-dimensional distribution.
\newblock \emph{Neural Computation}, 13\penalty0 (7):\penalty0 1443--1471, 2001.

\bibitem[Seliya et~al.(2021)Seliya, Abdollah~Zadeh, and Khoshgoftaar]{seliya2021literature}
N.~Seliya, A.~Abdollah~Zadeh, and T.~M. Khoshgoftaar.
\newblock A literature review on one-class classification and its potential applications in big data.
\newblock \emph{Journal of Big Data}, 8:\penalty0 1--31, 2021.

\bibitem[Shawe-Taylor and Cristianini(2004)]{shawe2004kernel}
J.~Shawe-Taylor and N.~Cristianini.
\newblock \emph{Kernel Methods for Pattern Analysis}.
\newblock Cambridge university press, 2004.

\bibitem[Suykens(2017)]{suykens2017deep}
J.~A. Suykens.
\newblock Deep restricted kernel machines using conjugate feature duality.
\newblock \emph{Neural Computation}, 29\penalty0 (8):\penalty0 2123--2163, 2017.

\bibitem[Suykens and Vandewalle(1999)]{suykens1999least}
J.~A. Suykens and J.~Vandewalle.
\newblock Least squares support vector machine classifiers.
\newblock \emph{Neural Processing Letters}, 9:\penalty0 293--300, 1999.

\bibitem[Tonin et~al.(2021)Tonin, Pandey, Patrinos, and Suykens]{tonin2021unsupervised}
F.~Tonin, A.~Pandey, P.~Patrinos, and J.~A. Suykens.
\newblock Unsupervised energy-based out-of-distribution detection using stiefel-restricted kernel machine.
\newblock In \emph{2021 International Joint Conference on Neural Networks (IJCNN)}, pages 1--8. IEEE, 2021.

\bibitem[Xiao et~al.(2024)Xiao, Pan, Liu, Zhao, Kong, and Hao]{xiao2024privileged}
Y.~Xiao, G.~Pan, B.~Liu, L.~Zhao, X.~Kong, and Z.~Hao.
\newblock Privileged multi-view one-class support vector machine.
\newblock \emph{Neurocomputing}, 572:\penalty0 127186, 2024.

\bibitem[Xing and He(2022)]{xing2022adaptive}
H.-J. Xing and Z.-C. He.
\newblock Adaptive loss function based least squares one-class support vector machine.
\newblock \emph{Pattern Recognition Letters}, 156:\penalty0 174--182, 2022.

\bibitem[Xing and Li(2020)]{xing2020robust}
H.-J. Xing and L.-F. Li.
\newblock Robust least squares one-class support vector machine.
\newblock \emph{Pattern Recognition Letters}, 138:\penalty0 571--578, 2020.

\bibitem[Yepes et~al.(2018)Yepes, L{\'o}pez, P{\'e}rez-Marcos, Gil, and Villarrubia]{yepes2018listen}
F.~A. Yepes, V.~F. L{\'o}pez, J.~P{\'e}rez-Marcos, A.~B. Gil, and G.~Villarrubia.
\newblock Listen to this: Music recommendation based on one-class support vector machine.
\newblock In \emph{Hybrid Artificial Intelligent Systems: 13th International Conference, HAIS 2018, Oviedo, Spain, June 20-22, 2018, Proceedings 13}, pages 467--478. Springer, 2018.

\bibitem[Zhao et~al.(2020)Zhao, Huang, Hu, and Li]{zhao2020improved}
Y.-P. Zhao, G.~Huang, Q.-K. Hu, and B.~Li.
\newblock An improved weighted one class support vector machine for turboshaft engine fault detection.
\newblock \emph{Engineering Applications of Artificial Intelligence}, 94:\penalty0 103796, 2020.

\bibitem[Zheng et~al.(2023)Zheng, Wang, and Chen]{zheng2023multikernel}
Y.~Zheng, S.~Wang, and B.~Chen.
\newblock Multikernel correntropy based robust least squares one-class support vector machine.
\newblock \emph{Neurocomputing}, 545:\penalty0 126324, 2023.

\end{thebibliography}
\bibliographystyle{abbrvnat}
\end{document}